%% file: main.tex
\documentclass[runningheads]{llncs}

\usepackage[mobile]{eccv}

\usepackage{eccvabbrv}

\input{premable}

\usepackage{graphicx}
\usepackage{booktabs}

\usepackage[accsupp]{axessibility}  

\usepackage[pagebackref,breaklinks,colorlinks,citecolor=eccvblue]{hyperref}

\usepackage{orcidlink}

\usepackage{makecell}

\usepackage{wrapfig}

\usepackage{tcolorbox}
\tcbuselibrary{listings,breakable}
\newtcolorbox{codebox}[1][]{
  colback=gray!10,
  colframe=black,
  boxrule=0.4pt,
  arc=2pt,
  top=2pt,
  bottom=2pt,
  left=3pt,
  right=3pt,
  boxsep=3pt,
  fontupper=\small,
  listing only,
  breakable,
  listing options={
    basicstyle=\ttfamily\footnotesize,
    language=Python,
    keywordstyle=\color{blue},
    stringstyle=\color{red},
    commentstyle=\color{gray},
    morekeywords={\quad}
  },
  #1
}

\newtcolorbox{widecodebox}[1][]{
  colback=gray!10,
  colframe=black,
  boxrule=0.4pt,
  arc=2pt,
  top=2pt,
  bottom=2pt,
  left=3pt,
  right=3pt,
  boxsep=3pt,
  fontupper=\small,
  listing only,
  listing options={
    basicstyle=\ttfamily\footnotesize,
    language=Python,
    keywordstyle=\color{blue},
    stringstyle=\color{red},
    commentstyle=\color{gray},
    morekeywords={\quad}
  },
  #1
}

\begin{document}
\title{CoMoVi: Co-Generation of 3D Human Motions and Realistic Videos} 


\author{
Chengfeng Zhao\inst{1} \and
Jiazhi Shu\inst{2} \and
Yubo Zhao\inst{1} \and
Tianyu Huang\inst{3} \and
Jiahao Lu\inst{1} \and \\
Zekai Gu\inst{1} \and
Chengwei Ren\inst{1} \and
Zhiyang Dou\inst{4} \and
Qing Shuai\inst{5} \and
Yuan Liu\inst{1}\textsuperscript{\Letter}
}


\institute{$^{1}$HKUST, $^{2}$SCUT, $^{3}$HKU, $^{4}$MIT, $^{5}$ZJU}

\maketitle

\vspace{-20pt}
\input{sec/0-abstract}    
\input{sec/1-introduction}
\input{sec/2-rw}
\input{sec/3-method}
\input{sec/4-exp}
\input{sec/5-conclusion}
\input{sec/X_suppl}

\newpage

%
%
\bibliographystyle{splncs04}
\bibliography{main}
\end{document}

%% file: premable.tex
\usepackage{graphicx}
\usepackage{amsmath}
\usepackage{amssymb}
\usepackage{booktabs}
\usepackage{bm}
\usepackage{physunits}

\usepackage{cuted}
\usepackage{capt-of}

\usepackage{multirow}
\usepackage{verbatim}
\usepackage{comment}
\usepackage{adjustbox}
\usepackage{wrapfig}
\usepackage{tikz}
\usepackage{stfloats} 
\usepackage{ulem}     

\usepackage{url}            
\usepackage{amsfonts}       
\usepackage{nicefrac}       
\usepackage{microtype}      

\usepackage{array}
\usepackage{xspace}
\usepackage{pifont}

\usepackage[dvipsnames]{xcolor}
\usepackage[table]{xcolor}
\usepackage[misc]{ifsym}

\usepackage[accsupp]{axessibility}

\makeatletter
\renewcommand{\maketag@@@}[1]{\hbox{\m@th\normalsize\normalfont#1}}%
\makeatother

%% file: sec/0-abstract.tex
\begin{abstract}
In this paper, we find that the generation of 3D human motions and 2D human videos is intrinsically coupled. 3D motions provide the structural prior for plausibility and consistency in videos, while pre-trained video models offer strong generalization capabilities for motions. Based on this, we present \textbf{CoMoVi}, a co-generative framework that generates 3D human motions and videos synchronously within a single diffusion denoising loop. However, since the 3D human motions and the 2D human-centric videos have a modality gap between each other, we propose to project the 3D human motion into an effective 2D human motion representation that effectively aligns with the 2D videos. Then, we design a dual-branch diffusion model to couple human motion and the video generation process with mutual feature interaction and 3D-2D cross attentions. To train and evaluate our model, we curate \textbf{CoMoVi-Dataset}, a large-scale real-world human video dataset with text and motion annotations, covering diverse and challenging human motions. Extensive experiments demonstrate that our method generates high-quality 3D human motion with a better generalization ability and that our method can generate high-quality human-centric videos without external motion references. Project page: \href{https://igl-hkust.github.io/CoMoVi/}{https://igl-hkust.github.io/CoMoVi/}.

\keywords{human motion synthesis \and human video generation \and 3D-2D co-generation \and dual-branch video diffusion architecture}

\end{abstract}

%% file: sec/1-introduction.tex
\section{Introduction}
\label{sec:intro}

The generation of 3D human motion and realistic video sequences is essential to understanding human behaviors and coherent visual dynamics, with broad downstream applications including character animation, VR/AR, and gaming. 

Traditional text-driven motion generation models learn to generate 3D human motion given textual descriptions~\cite{tevet2022mdm,zhang2023remodiffuse,chen2023mld,dai2024motionlcm,zhou2024emdm,meng2025mardm,ahn2018text2action,zhang2023t2mgpt,jiang2023motiongpt,guo2024momask,xiao2025motionstreamer,fan2025gotozero}. However, these approaches are often constrained by the bottleneck of high-quality 3D motion data, which leads to limited generalization capabilities and low prompt fidelity of these methods. 
Recently, advanced approaches~\cite{huang2025animax,pi2024motion2to3,lin2025vimogen} attempt to overcome these limitations by first generates human videos and then applying video-based motion capture algorithms to recover 3D human motion. While video generation models generalize well, they often struggle with highly structured objects such as human bodies, frequently producing implausible motions with inconsistent body structure, which in turn corrupts the recovered 3D motion.

\begin{figure*}[t!]
    \centering
    \includegraphics[width=\textwidth]{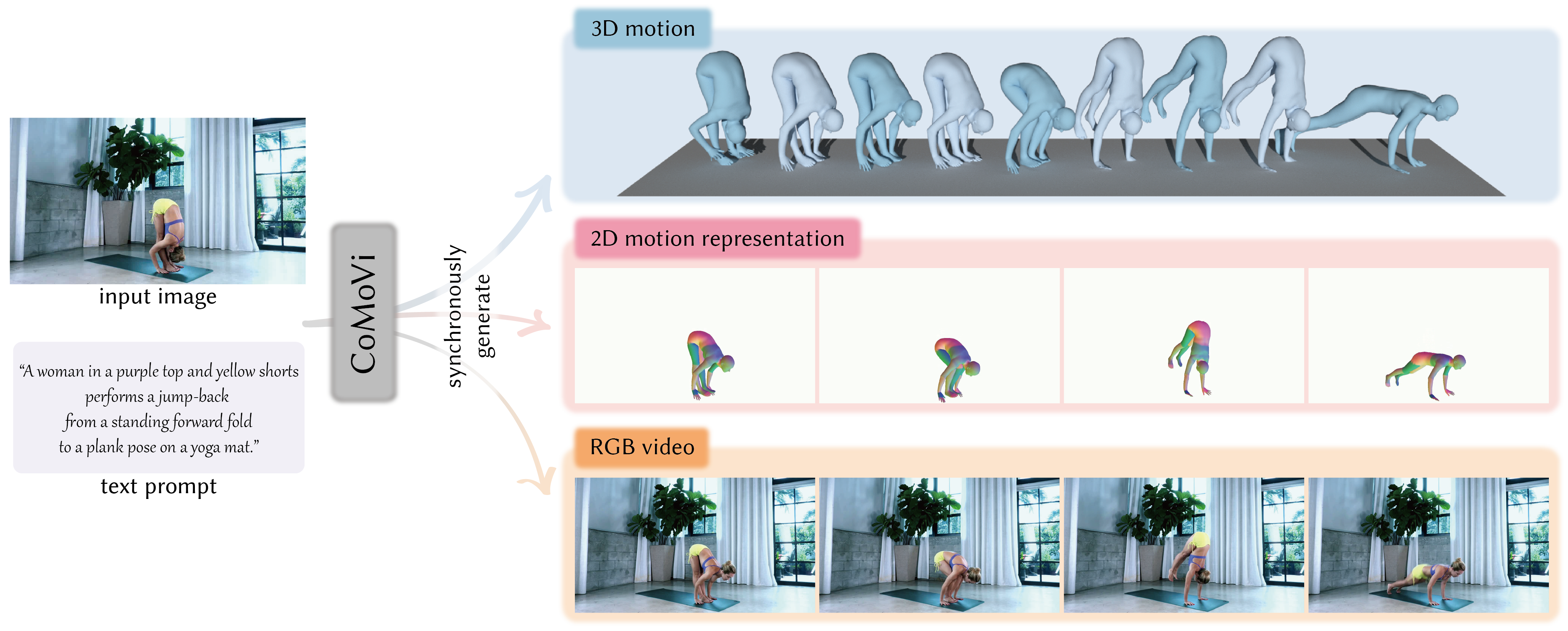}
    \vspace{-20pt}
    \caption{Given an input human image and motion description, \textbf{CoMoVi} generates 3D human motion and realistic video synchronously.}
    \label{fig:teaser}
    \vspace{-15pt}
\end{figure*}

On the other hand, although recent Video Diffusion Models (VDMs)~\cite{hong2022cogvideo,blattmann2023svd,yang2024cogvideox,wan2025wan,agarwal2025cosmos} demonstrate remarkable performance and strong generalization capabilities, the generation of a video about a specific person performing a particular action with high fidelity remains challenging. Previous human-centric video generation works introduce such priors by pre-extracting 2D pose or 3D motion as driving signals to guide the video generation process~\cite{hu2024animate,niu2024mofa,zhang2024mimicmotion,gan2025humandit,taghipour2025latentmove,zhu2024champ,zhou2024realisdance,zhou2025realisdance,he2025posegen}, achieving promising results with high quality. However, these methods all require external videos and motions as reference to provide such a guidance, and it remains unexplored how to generate high-quality video without such external video and motion references.

\begin{figure}[t!]
    \centering
    \includegraphics[width=\linewidth]{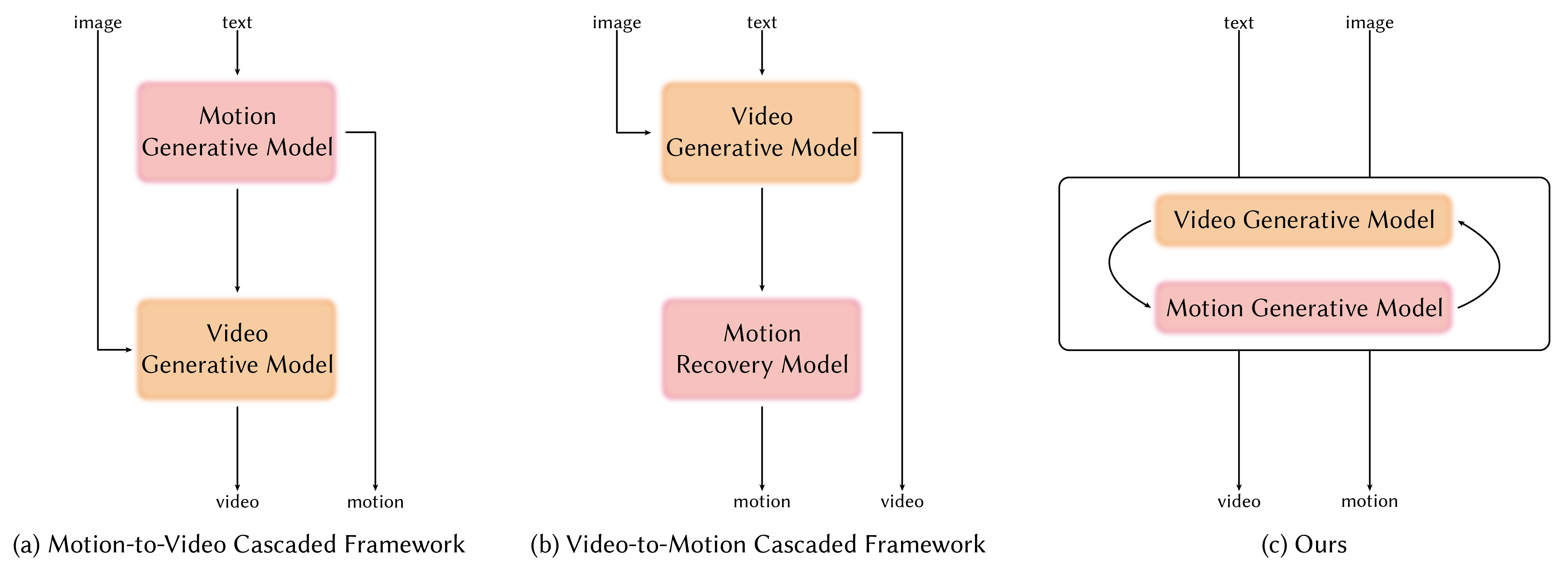}
    \vspace{-20pt}
    \caption{Different paradigms of motion video co-generation.}
    \label{fig:paradigm}
\vspace{-15pt}
\end{figure}

Through our discussion above, it is revealed that there exists a strong coupling relationship between 3D human motion and video generation. High-quality 3D motion can derive high-fidelity generated videos, and, conversely, the powerful prior of VDMs can enhance the generalization capabilities of 3D motion generation. However, as illustrated in Fig.~\ref{fig:paradigm}, existing works are cascaded, either in a motion-to-video or a video-to-motion framework, which are suboptimal. In this paper, we introduce \textbf{CoMoVi}, a novel framework for the co-generation of 3D human motion and human video synchronously. This co-generative framework allows mutual information exchange between the motion and video generation processes, enabling generalization enhancement for motion generation and structure guidance for video generation.

To achieve this synchronous co-generation, we first need to address the inherent modality gap between 3D human motion and 2D video. A natural approach to bridging this gap is to render 3D human motion into 2D feature maps, which allows for the synchronous generation of both motion and video utilizing VDMs. However, effectively formulating this 2D representation remains a critical challenge. Utilizing normal maps as the representation struggles to disambiguate symmetrical or overlapping body parts (e.g., distinguishing the left hand from the right) and lacks explicit structural semantics. Conversely, semantic maps provide part-level awareness, but inherently discard crucial 3D geometric information. To overcome these limitations, we propose a novel 2D human motion representation that successfully encapsulates both rich semantic distinctions and precise 3D structural geometry into the pixel space. As demonstrated in our experiments, this unified representation is pivotal to effectively aligning the two modalities and serves as the cornerstone of our co-generation framework.

Building upon this representation, we design a unified pipeline that takes an input image and a text description to synchronously generate human motions and the corresponding video sequences within a single diffusion denoising loop. Specifically, we develop a dual-branch VDM extended from Wan2.2-I2V-5B~\cite{wan2025wan}. This architecture tightly couples the denoising processes of 2D motion videos and RGB videos through mutual feature interactions, ensuring that the video generation is explicitly guided by robust human motion priors and that the motion generation shares the generalization ability of video generation. 
Furthermore, instead of solving for a 3D motion from our generated 2D motion representations with optimization, we design a module to combine the features from both videos and 2D motion maps for direct 3D motion estimation. Thus, CoMoVi can directly output both 3D human motions and 2D human-centric videos.

Finally, effective training and evaluation of our co-generation framework inherently demand a dataset that simultaneously provides comprehensive text descriptions, high-quality video sequences, and precise 3D motion annotations. Existing datasets often fall short of these stringent requirements: while large-scale datasets like Motion-X++~\cite{lin2023motionx} suffer from compromised video quality and low resolution, high-fidelity datasets such as HumanVid~\cite{wang2024humanvid} are limited in scale and rich 3D annotations, with less than 20k real-world videos. To bridge this critical data gap, we contribute the CoMoVi-Dataset. Leveraging a meticulously designed data processing pipeline, we curate a collection of approximately 50K high-resolution, real-world human videos. Our dataset significantly surpasses HumanVid~\cite{wang2024humanvid} in scale while offering vastly superior video quality compared to Motion-X++~\cite{lin2023motionx}. Furnished with reliable text and motion annotations, the CoMoVi-Dataset covers a diverse array of challenging human dynamics, serving as a robust foundation for our model and future research in this domain.

To demonstrate the effectiveness of CoMoVi, we conduct comprehensive experiments on the Motion-X++ dataset~\cite{lin2023motionx}, VBench benchmark~\cite{huang2023vbench,huang2024vbench++,zheng2025vbench2}, and our CoMoVi-Dataset for both motion generation and video generation tasks. Qualitative and quantitative results validate that our approach is effective in generalizable 3D human motion generation, outperforming state-of-the-art (SoTA) text-to-motion (T2M). For human-centric video generation, CoMoVi achieves high-quality video generation without relying on external reference videos and pre-extracted motion signals.

%% file: sec/2-rw.tex
\section{Related Works}
\label{sec:rw}

\paragraph{\bf Text-driven Human Motion Synthesis.} Along with the emergence of large-scale human motion datasets with natural language labels~\cite{plappert2016kit,mahmood2019amass,punnakkal2021babel,guo2022ml3d,black2023bedlam,lin2023motionx,wang2024quo,lu2025scamo,fan2025gotozero,ren2025motionpro,wu2025finemotion,wang2024beingbyondm0,cao2025beingbeyondm05,tesch2025bedlam2,mclean2025embody,cao2025reconstructing,dou2023tore,wu2024dice,cui2025user}, T2M methods step a lot forward in both diffusion~\cite{tevet2022mdm,zhang2023remodiffuse,chen2023mld,dai2024motionlcm,zhou2024emdm,meng2025mardm} and autoregressive  directions~\cite{ahn2018text2action,zhang2023t2mgpt,jiang2023motiongpt,guo2024momask,wu2024motion,xiao2025motionstreamer}. However, the scarcity of high-quality 3D motion data constraints their diversity and generalization capabilities. To overcome this limitation, recent works leverage multi-view diffusion models~\cite{liu2023syncdreamer,huang2025mvadapter,cai2025up2you,chen2025synchuman} to generate 2D motion sequences and then triangulate to 3D~\cite{pi2024motion2to3,huang2025animax}. Though effective, these approaches separate 2D generation and 3D reconstruction into two independent processes and only consider human motion as 2D joint coordinates, neglecting the coupling relationship between 3D motion and 2D frames. Our work proposes to encode 3D motion into the same space pre-trained VDMs~\cite{hong2022cogvideo,blattmann2023svd,yang2024cogvideox,kong2024hunyuanvideo,wan2025wan,agarwal2025cosmos} to generate 3D motion and 2D video synchronously.

\vspace{-2mm}
\paragraph{\bf Image-based Human Animation.} Based on powerful image generation, pioneering image animation works animate static images by fine-tuning or adding additional motion adapters~\cite{guo2023animatediff,karras2023dreampose} to pre-trained text-to-image (T2I) models~\cite{rombach2022sd}. Inspired by the success of ControlNet~\cite{zhang2023controlnet}, a series of motion-driven methods are developed, utilizing 2D pose~\cite{yang2023dwpose} to drive human video generation~\cite{hu2024animate,niu2024mofa,zhang2024mimicmotion,gan2025humandit,taghipour2025latentmove}. Recent works incorporate more multimodal control signals such as 3D parametric body model~\cite{loper2023smpl,romero2022mano,pavlakos2019smplx,zhu2024champ,zhou2024realisdance,zhou2025realisdance,he2025posegen}, camera trajectory~\cite{wang2024humanvid,li2025tokenmotion,shao2025interspatial}, optical flow~\cite{shi2024motioni2v,liao2025motionagent}, 3D scene geometry~\cite{cao2025uni3c,gu2025das}, novel background~\cite{pan2024actanywhere,liang2025realismotion,li2025humangenesis,niu2025anicrafter}, and audio~\cite{huang2025homa,chen2025humo,wang2025interacthuman} to empower multi-subject interaction animation~\cite{wang2025multianimate,wang2025dreamactor}, video subject replacement~\cite{cheng2025wanimate}, and promptable animation~\cite{kim2025tavid,jin2025matrix}. Additionally, researchers also explore how to implicitly transfer high-level motion patterns directly from reference to target videos~\cite{tan2024animatex,tan2025animatepp,song2025xunimotion,bian2025videoasprompt}, bypassing the extraction of explicit driving motion signals. Yet, such methods still require extra reference videos to drive the animation and are not capable of co-generating motions and videos.

\vspace{-2mm}
\paragraph{\bf Joint Generation of Human Motion and Video.} In order to achieve co-generation and remove the dependency of driving sources, latest works follow a cascaded generation pipeline by tailoring one generative model to the other. Motion-to-video methods generate human videos conditioning on pre-generated 2D pose~\cite{gan2025humandit,wang2025humandreamer,xi2025toward}, 3D motion~\cite{liu2025ponimator,nam2025cameo,li2025genhsi,wang2025mosa,huang2025movein2d} or optical flow~\cite{liao2025motionagent} sequences, and video-to-motion frameworks~\cite{liu2025revision} first generate human video and then refine it using the motion estimation results. Nevertheless, cascaded pipelines propagate defects in the upstream model and overlook the coupling relationship of two generative processes. In the field of multimodal generation and understanding, advanced approaches co-generate RGB videos and normals~\cite{xi2025omnivdiff,wang2025multianimate,wang2025mmgen}, depths~\cite{xi2025omnivdiff,wang2025multianimate,wang2025mmgen,xiong2025huprior3r,lu2025align3r,lu2025trackingworld,li2026unish,lu2026track4world}, segmentations~\cite{xi2025omnivdiff,wang2025mmgen}, optical flow~\cite{chefer2025videojam}, and motion maps~\cite{lei2025momaps}. However, the co-generation of 3D human motions and 2D human videos has not been well studied, and the few concurrent works~\cite{pang2025unimo,yang2025echomotion,lin2025vimogen} also reveal the promising prospects of this research direction.

%% file: sec/3-method.tex
\section{Method}
\label{sec:method}

\subsection{Overview}
\begin{figure*}[t!]
    \centering
    \includegraphics[width=\textwidth]{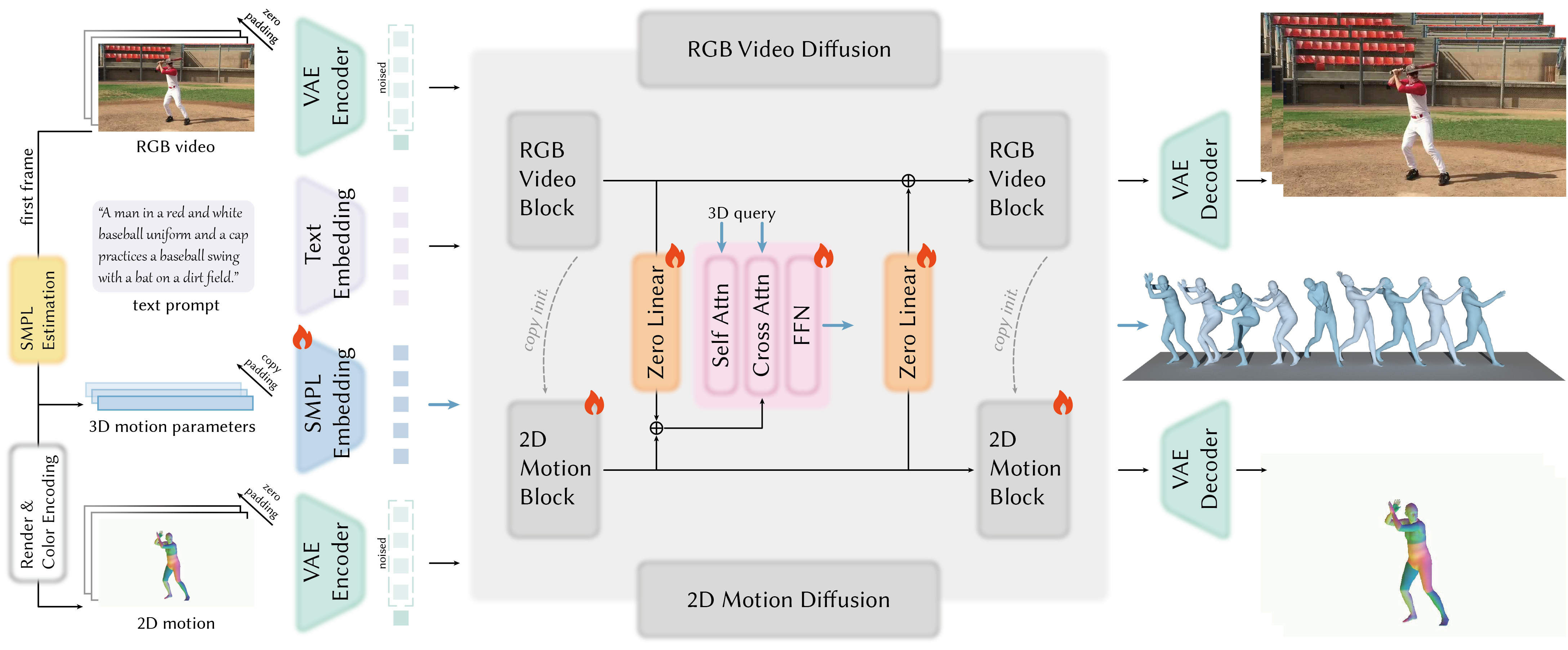}
    \vspace{-20pt}
    \caption{Pipeline overview of CoMoVi. Our method consists of an effective 2D human motion representation (Sec.~\ref{sec:morep}) to encode 3D motion information in pixel space, and a dual-branch diffusion model extended from Wan2.2-I2V-5B to coordinate 2D motion and RGB video sequence denoising process with 3D-2D cross-attention modules to concurrently generate 3D human motion (Sec.~\ref{sec:comovi}).}
    \label{fig:pipeline}
\vspace{-15pt}
\end{figure*}

Our goal is to co-generate 3D human motion $\left\{\bm{m}_i\in \mathbb{R}^{J\times 3}\right\}_{i=0}^F$ and RGB video $\left\{\bm{s}_i\in \mathbb{R}^{H\times W\times 3}\right\}_{i=0}^F$ sequence in $F$ frames given a starting image $\bm{s}_0$ and a text description $\bm{p}$. In which $J$ is the number of body joints defined by SMPL~\cite{loper2023smpl}, and $H\times W$ is the resolution of generated videos. As illustrated in Fig.~\ref{fig:pipeline}, we first estimate the 3D human motion $\bm{m}_0$ of $\bm{s}_0$ using CameraHMR~\cite{patel2025camerahmr} and render the 3D SMPL mesh posed in $\bm{m}_0$ as our 2D human motion representation $\bm{k}_0\in \mathbb{R}^{H\times W\times 3}$ according to vertex normals and body part semantics (Sec.~\ref{sec:morep}). Then, $\bm{s}_0$ and $\bm{k}_0$ are zero-padded to $F$ frames and fed into each branch of our dual-branch diffusion model, together with $\bm{m}_0$ and $\bm{p}$ to generate human video $\left\{\bm{s}_i\right\}_{i=0}^F$, 2D motion map $\left\{\bm{k}_i\right\}_{i=0}^F$, and 3D human motion $\left\{\bm{m}_i\right\}_{i=0}^F$ sequence synchronously (Sec.~\ref{sec:comovi}). Our training data is prepared as introduced by Sec.~\ref{sec:dataset}.

\subsection{2D Human Motion Representation}
\label{sec:morep}

In this section, we aim to design a 2D motion representation composed of normal and semantic maps for the co-generation, which not only carries rich 3D information but also aligns with the 2D video. Existing works~\cite{zhu2024champ} treat normal and semantic maps as separate conditions. This leads to inherent limitations: normal maps lack the semantic information necessary to distinguish between different body parts, while semantic maps are devoid of 3D structural cues. To address this limitation, we propose an improved motion representation that cohesively integrates both types of information.

\begin{wrapfigure}{r}{0.42\textwidth}
    \centering
    \vspace{-25pt}
    \includegraphics[width=0.4\textwidth]{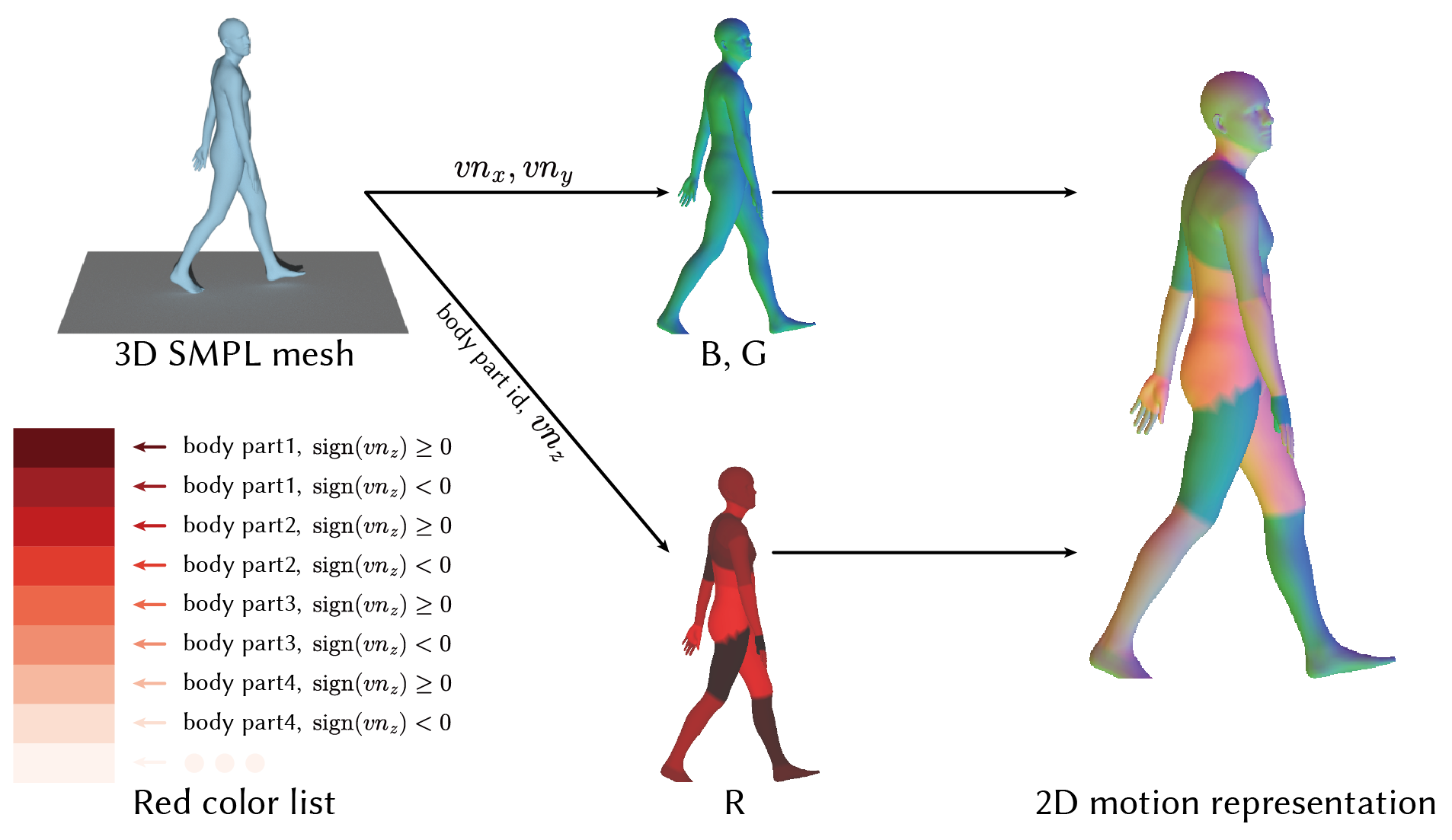}
    \vspace{-10pt}
    \caption{We integrate normals and body part semantics of 3D SMPL meshes into a single representation.}
    \label{fig:representation}
    \vspace{-20pt}
\end{wrapfigure}
To be specific, given the $i$-th vertex $\bm{ve}_i=\left(ve_x,ve_y,ve_z\right)$ of SMPL mesh, the vertex normal $\bm{vn}_i=\left(vn_x,vn_y,vn_z\right)$ satisfies
\begin{equation}
    vn_z = \pm\sqrt{1-vn_x^2-vn_y^2},
\end{equation}
providing that $vn_x$ and $vn_y$ are known and only $\operatorname{sign}(vn_z)$ is undetermined. Thus, we can combine this sign with body part semantics and encode both into a single vertex color channel. As illustrated in Fig.~\ref{fig:representation}, we first encode $vn_x$ and $vn_y$ as the Blue and Green channels, respectively. Then, suppose that SMPL can be segmented into $R$ body parts, we define a color list containing $2R$ candidate values uniformly sampled in the range $\left[0,1\right]$ for the Red channel. The Red channel value for $\bm{ve}_i$ of part $r$ is assigned as
\begin{equation}
    \mathrm{Red}(\bm{ve}_i) =
    \begin{cases}
        \mathrm{RedList[r]} & \text{if } \operatorname{sign}(vn_z) \geq 0 \\
        \mathrm{RedList[r+1]} & \text{if } \operatorname{sign}(vn_z) < 0\ .
    \end{cases}
\end{equation}
This strategy enables effective compression of 3D surface normals and body part semantics into a single RGB image, which preserves essential 3D structural information and can also be embedded in the latent space of VDMs seamlessly.

\subsection{Co-Generation of Human Motion and Video}
\label{sec:comovi}

To achieve co-generation of human motion and video, we adopt a dual-branch diffusion architecture and train it in multiple stages progressively. In the following, we use $\bm{x}_t^\text{motion}$ and $\bm{x}_t^\text{video}$ to represent the latents of $\mathcal{D^{\text{motion}}}$ and $\mathcal{D^{\text{video}}}$ at noise level $t$, and the clean latents are $\bm{x}_0^\text{motion}$ and $\bm{x}_0^\text{video}$ respectively.

\vspace{-2mm}
\paragraph{\bf Adapt Pre-trained VDM to 2D Motion Domain.}
\begin{wrapfigure}{r}{0.42\textwidth}
    \centering
    \vspace{-20pt}
    \includegraphics[width=0.4\textwidth]{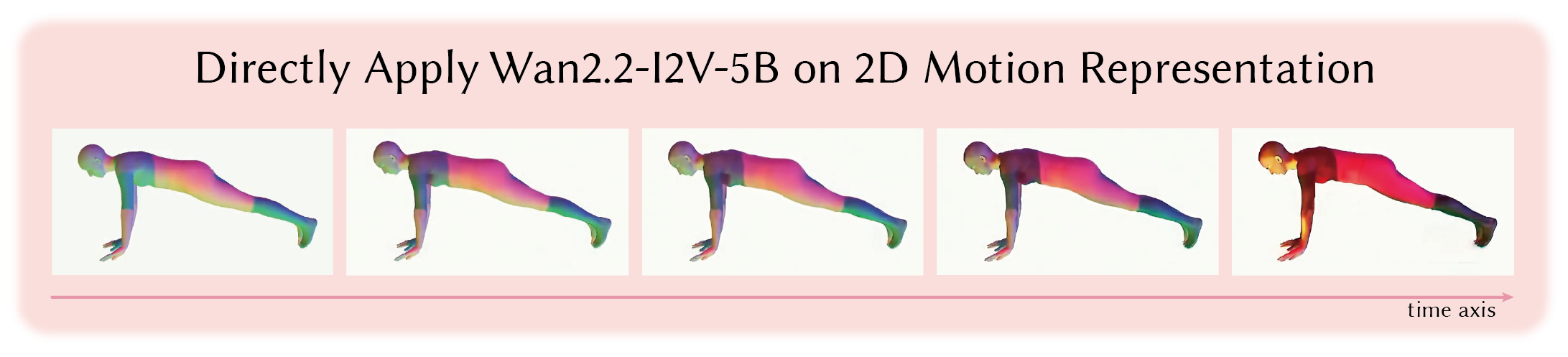}
    \vspace{-10pt}
    \caption{We observe that significant appearance shifts occur when directly applying pretrained VDM on our 2D motion representation.}
    \label{fig:appearance_shift}
    \vspace{-15pt}
\end{wrapfigure}

Our first training stage is to finetune $\mathcal{D^{\text{motion}}}$ to adapt its weights from the RGB video to our 2D motion representation domain. This is necessary because significant appearance shifts occur when directly applying the pre-trained VDM to our 2D motion representation sequence (see Fig.~\ref{fig:appearance_shift}), corrupting the essential color patterns that encompass rich 3D information. In this stage, only $\mathcal{D^{\text{motion}}}$ is tunable, which learns to denoise $\bm{x}_t^\text{motion}$ to $\bm{x}_0^\text{motion}$ conditioning on the text prompt and the starting frame.

\vspace{-2mm}
\paragraph{\bf Couple Motion and Video Generation Process.} In the second training stage, we introduce mutual feature interactions between dual branches to guide human video generation, and 3D-2D cross-attention module to output 3D human motion from video latents. Only $\mathcal{D^{\text{video}}}$ is freezed in this stage.

\vspace{-2mm}
\paragraph{Mutual Feature Interactions.} For the $i$-th diffusion block, we insert a group of two zero-linear modules after it between dual branches to obtain
\begin{align}
    \bm{x}_t^\text{fused} &= \bm{x}_t^\text{motion} + \operatorname{ZeroLinear}_{i}(\bm{x}_t^\text{video}) \notag \\
    \bm{x}_t^\text{video} &= \bm{x}_t^\text{video} + \operatorname{ZeroLinear}_{i+1}(\bm{x}_t^\text{motion}).
\label{eq:fuse}
\end{align}
Then, we pass $\bm{x}_t^\text{motion}$ and $\bm{x}_t^\text{video}$ to the next $(i+1)$-th diffusion block of $\mathcal{D}^{\text{motion}}$ and $\mathcal{D}^{\text{video}}$ respectively. The fused latent $\bm{x}_t^\text{fused}$ will serve as the key and value in our 3D-2D cross-attention module.

\vspace{-2mm}
\paragraph{3D-2D Cross-Attention Module.} To generate 3D human motion synchronously, we propose a 3D-2D cross-attention module $\mathcal{A}$ composed of 6 layers of self-attention, cross-attention, and feed-forward network combination. We first initialize all $F$-frame SMPL poses with the known initial pose $\bm{m}_0$, and apply a linear SMPL embedding layer to project them into $F$-frame embeddings $\bm{q}\in\mathbb{R}^{F\times C}$. Subsequently, these initial pose embeddings are used as 3D queries going through self-attention layers as
\begin{equation}
    \bm{q} = \operatorname{SelfAttention}(\bm{q}).
\end{equation}
Then $\bm{q}$ will interact with $\bm{x}_t^{\text{fused}}$ to generate human poses for all frames.
Since $\bm{x}_t^{\text{fused}}$ is derived from 2D video latents which are compressed along temporal axis by the VAE encoder $\mathcal{E}$ with compression ratio as 4~\cite{wan2025wan}, 1 frame of $\bm{x}_t^{\text{fused}}$ corresponds to 4 frames of $\bm{q}$. Therefore, we re-organize $\bm{q}$ per 4 frames to get $\bm{q}'\in\mathbb{R}^{\frac{F-1}{4}\times4\times C}$ and apply 3D-2D cross attention as
\begin{equation}
    \bm{q} = \operatorname{CrossAttention}(\bm{q}',\bm{x}_t^{\text{fused}}).
\end{equation}
Finally, our model generates 3D human motion $\left\{\bm{m}_i\right\}_{i=1}^F$ by a feed-forward network and a final output projection layer to map $\bm{q}$ back to SMPL parametric dimension.

\vspace{-2mm}
\paragraph{\bf Losses.}
Our total training loss is defined as
\begin{equation}
    \mathcal{L}^{\text{total}} = \mathcal{L}^{\text{motion}} + \mathcal{L}^{\text{video}} + \mathcal{L}^{\text{smpl}},
\end{equation}
where $\mathcal{L}^{\text{motion}}$ and $\mathcal{L}^{\text{video}}$ have the same flow matching objective formulation~\cite{lipman2022flowmatching}
\begin{equation}
    \mathcal{L} = \mathbb{E}_{\bm{x}_0,\epsilon,t,\bm{p}}\left[ \left\Vert \mathcal{D}(\bm{x}_t,t,\bm{p}) - \bm{v}_t \right\Vert_2^2 \right],
\end{equation}
and $\mathcal{L}^{\text{smpl}}$ is defined as
\begin{equation}
    \mathcal{L}^{\text{smpl}} = \frac{1}{F-1}\sum_{i=1}^{F-1}\left\Vert\bm{m}_i-\operatorname{GT}(\bm{m}_i)\right\Vert_2^2.
\end{equation}
During training, the gradients from $\mathcal{L}^{\text{smpl}}$ backpropagate to the 2D motion branch $\mathcal{D}^{\text{motion}}$ and eventually update the information injected to RGB video branch $\mathcal{D}^{\text{rgb}}$. While in the inference phase, the primary function of $\mathcal{A}$ is to generate 3D human motion from the latents of VDM, which does not iteratively guide the denoising process.

\begin{table*}[t!]
    \centering
    \scriptsize
    \setlength{\tabcolsep}{22.2pt}
    \begin{tabular}{lccc}
        \toprule
        Dataset & Clips & Motion Type & Resolution\\
        \midrule
        Motion-X++~\cite{lin2023motionx} & 120,500 & 3D SMPL & -\\
        HumanVid*~\cite{wang2024humanvid} & 19,688 & 2D Pose & 1080P+\\
        MotionVid~\cite{wang2025humandreamer} & 1.2M & 2D Pose & -\\
        \cellcolor{lightgray!30}CoMoVi-Dataset & \cellcolor{lightgray!30}54,053 & \cellcolor{lightgray!30}3D SMPL & \cellcolor{lightgray!30}720P+\\
        \bottomrule
    \end{tabular}
    \vspace{2pt}
    \caption{Comparison of CoMoVi-Dataset with existing datasets. ``*": We only count real-world data.}
    \label{tab:dataset}
    \vspace{-20pt}
\end{table*}

\subsection{CoMoVi-Dataset}
\label{sec:dataset}

Training our co-generative framework requires a high-quality dataset of triplets containing human videos, 3D human motions, and text annotations. Different from previous datasets that focus on single-modality taks such as motion-only or video-only generation, we collect real-world videos paired with 3D motion for the co-generation task. Concurrent works~\cite{lin2025vimogen,yang2025echomotion} also make efforts and reveal the promising prospects of this research direction. As reported in Tab.~\ref{tab:dataset}, existing datasets~\cite{lin2023motionx,wang2024humanvid,wang2025humandreamer,lin2025vimogen,ren2023lip,zhao2024imhoi} fail to simultaneously meet our demands for high-quality video and sufficient 3D motion. Therefore, we curate such a dataset called CoMoVi-Dataset. Specifically, we source high-resolution human videos from Koala-36M~\cite{wang2025koala}, HumanVid~\cite{wang2024humanvid}, and publicly available Internet videos, and employ a carefully designed filtering pipeline to select clips featuring single-person motion based on Qwen3~\cite{qwen3}, Qwen2.5-VL~\cite{qwen2.5} and YOLO~\cite{redmon2016yolo}. For 3D human motion annotations, we obtain pseudo-labels using SoTA in-camera human mesh recovery method CameraHMR~\cite{patel2025camerahmr} for the best re-projection alignment. A smoothing post-processing procedure is tailored to further refine the frame-wise annotations. For text descriptions, we utilize the advanced Gemini-2.5-Pro to generate precise motion captions for each video. More details about our data filtering hierarchy and prompts are provided in the supplementary material.

%% file: sec/4-exp.tex
\section{Experiments}
\label{sec:experiments}

In this section, we first introduce our implementation and evaluation details. Then, we compare our approach with SoTA T2M and I2V models for motion generation and video generation tasks, respectively. Finally, we conduct ablation studies to validate the effectiveness of our proposed motion representation and model architecture designs.

\subsection{Implementation Details}
\label{sec:imp_details}

We train our model on the training set of our proposed dataset using 24 A100-SXM4-40G GPUs with per GPU batch size 1 and gradient accumulation steps 4 for 6,000 optimization steps. We use the ZeRO-3~\cite{rajbhandari2020zero} strategy and AdamW optimizer with learning rate 2e-5. Our training data are unified to resolution $H\times W=704\times1280$, $F=81$ frames in 16 fps. More implementation details can be found in the supplementary material.

\subsection{Evaluation Datasets and Metrics}
\label{sec:eval_metrics}
For comprehensive evaluations on both motion generation and video generation tasks, we take the widely-used Motion-X++ dataset~\cite{lin2023motionx}, VBench benchmark~\cite{huang2023vbench,huang2024vbench++,zheng2025vbench2}, and also our testing set to assess performance of all compared models. For the motion generation task, we adhere to the evaluation protocol defined by MoMask~\cite{guo2024momask} and use Frechet Inception Distance (FID), R-Precision (@1 and @3), and MultiModal Distance (MMDist) metrics. Since our method generates motions and videos from an input image and a text prompt, we adopt the Motion-X++ benchmark~\cite{lin2023motionx} containing texts, images, and 3D human motions. Though HumanML3D~\cite{guo2022ml3d} is widely used in motion generation evaluation, it lacks images, so not a good fit with our method. For the video generation task, we calculate ``Subject Consistency (SC)'', ``Background Consistency (BC)'', ``Motion Smoothness (MS)'', ``Aesthetic Quality (AQ)'', and ``Imaging Quality (IQ)'' metrics provided by VBench~\cite{huang2023vbench,huang2024vbench++,zheng2025vbench2}. The ``Dynamic Degree (DD)'' is excluded since it is aimed at measuring scene-level dynamics, on our testing set. For the following tables, the numbers marked in \textbf{bold} and \underline{underlined} represent the first and the second best, respectively.

\subsection{Comparisons on Motion Generation Task}
\label{sec:comp_mogen}

\paragraph{\bf Baselines.}
We compare with SoTA T2M methods, including diffusion-based MDM~\cite{tevet2022mdm} and autoregressive MotionGPT~\cite{jiang2023motiongpt}, Momask~\cite{guo2024momask}, and Go-to-Zero~\cite{fan2025gotozero}. Additionally, since we explore a new task of co-generating 3D human motions and videos from text prompts together with an input image, there does not exist a T2M baseline that perfectly matches our setting currently. Therefore, we build and compare with a simple yet meaningful baseline Wan2.2-I2V-5B+CameraHMR~\cite{wan2025wan,patel2025camerahmr} that first generates human video from a starting image and a text prompt, and then captures the 3D motion of the person in that generated video. It is noteworthy that we sample test sequences from Motion-X++~\cite{lin2023motionx} on our own since no official train/test split is provided, which might be included in the training set of Go-to-Zero~\cite{fan2025gotozero}.

\begin{figure*}[t!]
    \centering
    \includegraphics[width=\linewidth]{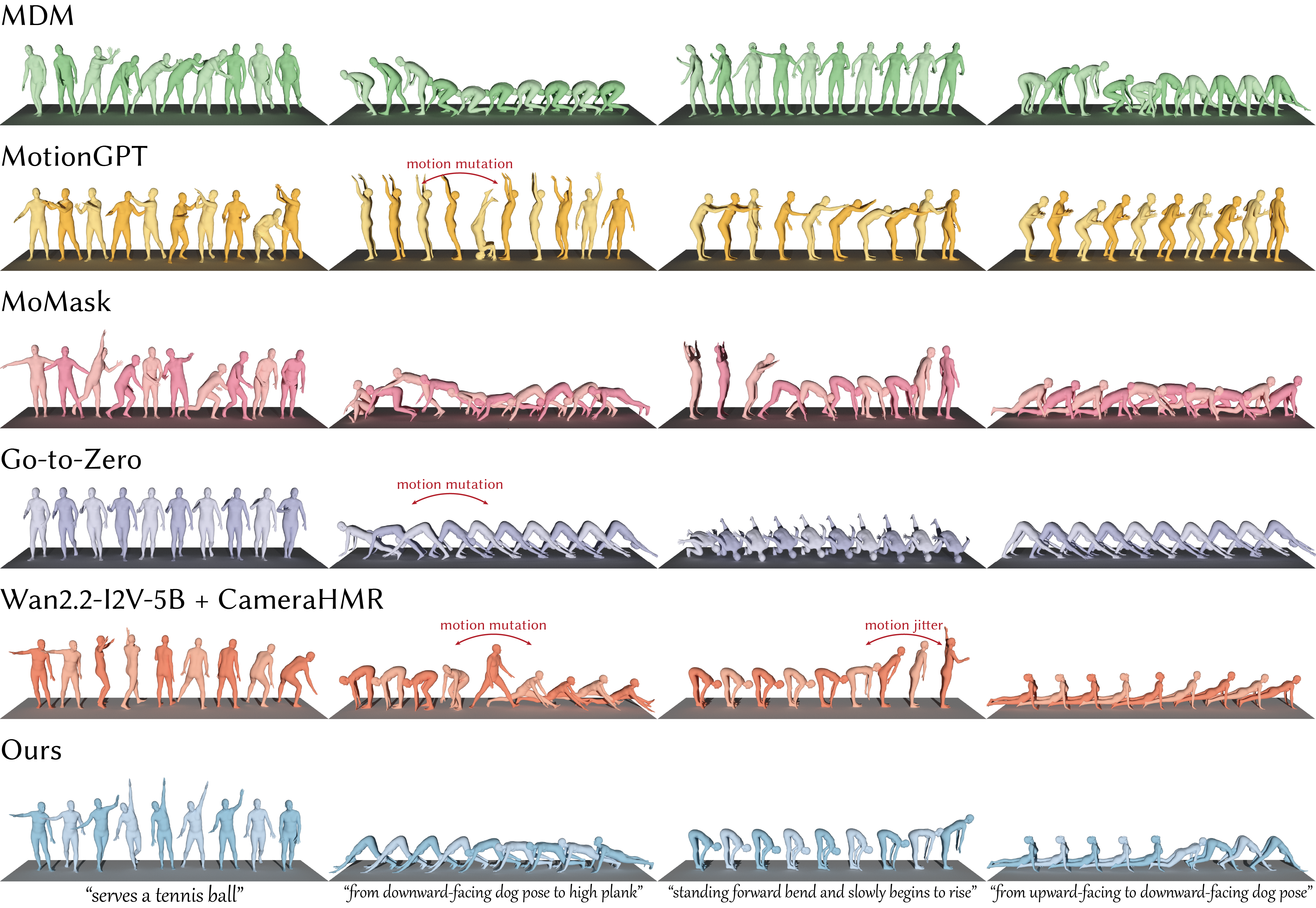}
    \vspace{-20pt}
    \caption{Qualitataive comparison of 3D human motion generation with SoTA T2M models~\cite{tevet2022mdm,jiang2023motiongpt,guo2024momask,fan2025gotozero}. Wan2.2-I2V-5B+CameraHMR~\cite{wan2025wan,patel2025camerahmr} is a simple yet meaningful baseline composed a video generation model followed by a video motion capture model. We present motion keywords in text prompts for simplicity.}
    \label{fig:mogen_comp}
\vspace{-5pt}
\end{figure*}

\vspace{-2mm}
\paragraph{\bf Results.}
As presented in Fig.~\ref{fig:mogen_comp}, CoMoVi generates 3D human motions with high prompt fidelity and dynamic smoothness, while baselines often produce jittery motions, unrelated contents, and implausible body movements. Quantitative evaluations in Tab.~\ref{tab:mogen} also validate that our co-generative framework outperforms SoTA T2M models on the testing set of CoMoVi-Dataset, and also generalizes well, achieving comparable performance on the unseen MotionX++ dataset~\cite{lin2023motionx} to Go-to-Zero~\cite{fan2025gotozero}. According to the results of Wan+CameraHMR~\cite{wan2025wan,patel2025camerahmr}, incorporating additional reference images helps achieve promising performance by providing structural cues. Advancing more, our co-generative framework effectively couples motion and video modalities, enabling them to mutually guide one another during the denoising process, leading to further improvements in both generated motion and video.

\begin{table*}[t!]
    \centering
    \scriptsize
    \setlength{\tabcolsep}{2.4pt}
    \begin{tabular}{lcccccccc}
        \toprule
        & \multicolumn{4}{c}{CoMoVi-Dataset} & \multicolumn{4}{c}{Motion-X++~\cite{lin2023motionx}}\\
        \cmidrule(lr){2-5} \cmidrule(lr){6-9}
        Method & FID $\downarrow$ & R@1 $\uparrow$ & R@3 $\uparrow$ & MMDist $\downarrow$ & FID $\downarrow$ & R@1 $\uparrow$ & R@3 $\uparrow$ & MMDist $\downarrow$ \\
        \midrule
        MDM~\cite{tevet2022mdm}                                 & 27.740 & 0.287 & 0.540 & 4.848 & 20.871 & 0.332 & 0.406 & 4.803 \\
        MotionGPT~\cite{jiang2023motiongpt}                     & 26.079 & 0.316 & 0.569 & 4.486 & 27.084 & 0.305 & 0.529 & 5.276 \\
        MoMask~\cite{guo2024momask}                             & 24.521 & 0.385 & 0.606 & 4.943 & 19.365 & 0.316 & 0.569 & 4.915 \\
        Go-to-Zero-3B*~\cite{fan2025gotozero}                   & 3.175  & 0.440 & 0.605 & 3.585 & \underline{15.474} & 0.311 & \underline{0.692} & 3.751 \\
        Go-to-Zero-7B*~\cite{fan2025gotozero}                   & 1.641  & 0.459 & \underline{0.608} & 3.429 & \textbf{14.470} & \textbf{0.384} & \textbf{0.699} & \underline{3.578} \\
        Wan+CameraHMR~\cite{wan2025wan,patel2025camerahmr}      & \underline{0.815} & \underline{0.539} & 0.591 & \underline{3.056} & 21.538 & 0.302 & 0.464 & 3.735 \\
        \cellcolor{lightgray!30}Ours                                                    & \cellcolor{lightgray!30}\textbf{0.349} & \cellcolor{lightgray!30}\textbf{0.565} & \cellcolor{lightgray!30}\textbf{0.640} & \cellcolor{lightgray!30}\textbf{3.035} & \cellcolor{lightgray!30}16.728 & \cellcolor{lightgray!30}\underline{0.377} & \cellcolor{lightgray!30}0.571 & \cellcolor{lightgray!30}\textbf{3.507} \\
        \bottomrule
    \end{tabular}
    \vspace{2pt}
    \caption{Quantitative evaluation of 3D human motion generation. ``*'': Motion-X++~\cite{lin2023motionx} is in the training set of Go-to-Zero~\cite{fan2025gotozero}.}
    \label{tab:mogen}
\vspace{-25pt}
\end{table*}

\begin{wrapfigure}{r}{0.42\textwidth}
    \centering
    \vspace{-50pt}
    \includegraphics[width=0.35\textwidth]{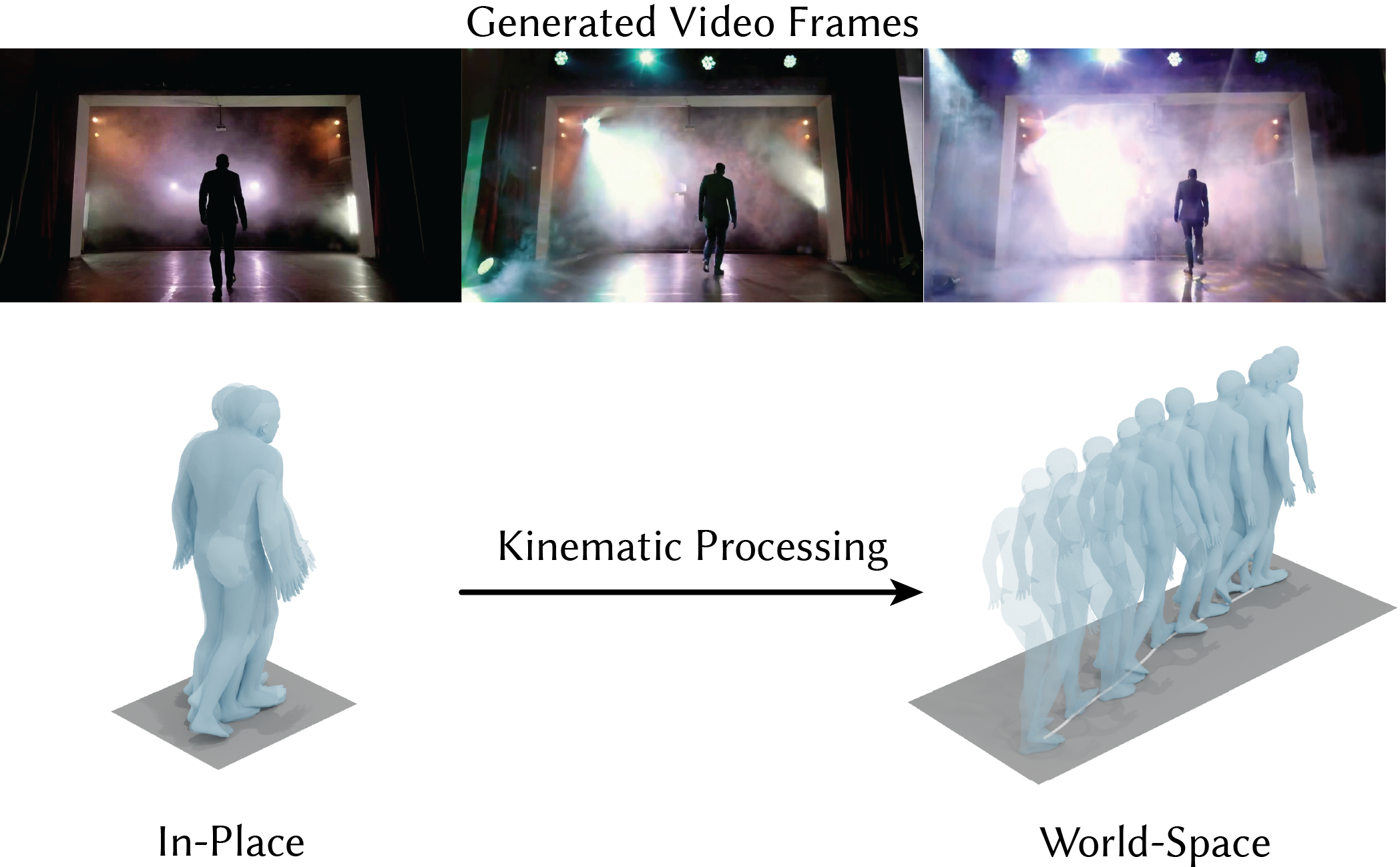}
    \vspace{-10pt}
    \caption{Extending CoMoVi results to world space via kinematic processing.}
    \label{fig:worldspace}
    \vspace{-30pt}
\end{wrapfigure}

\begin{figure*}[t!]
    \centering
    \includegraphics[width=\linewidth]{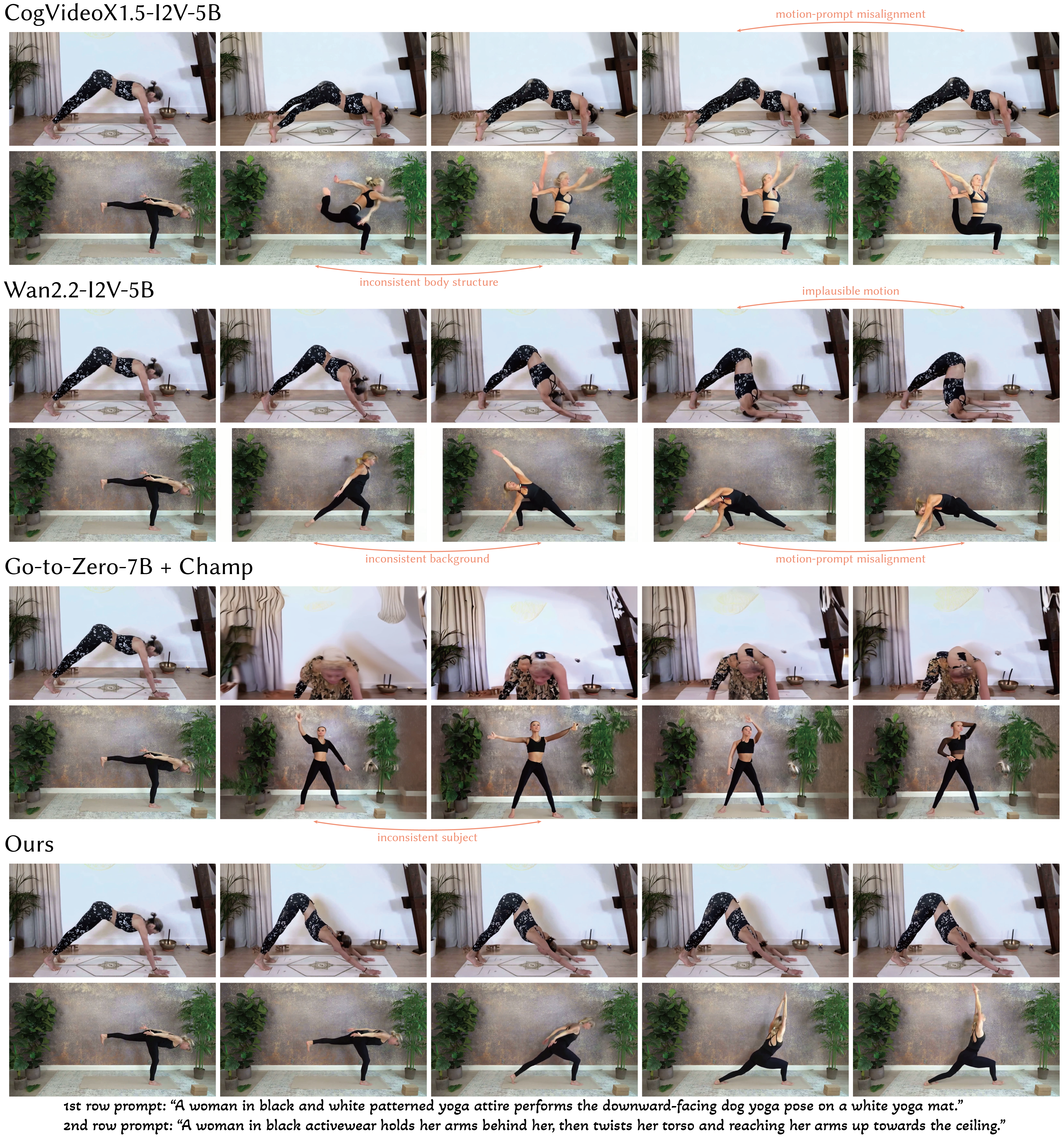}
    \vspace{-20pt}
    \caption{Qualitataive comparison of human video generation with SoTA open-souce I2V models~\cite{yang2024cogvideox,wan2025wan}, and a baseline composed of SoTA T2M~\cite{fan2025gotozero} and motion-driven video generation model~\cite{zhu2024champ}.}
    \label{fig:vigen_comp}
\vspace{-5pt}
\end{figure*}

\vspace{-2mm}
\paragraph{\bf Extending to World-Centric Motion Generation.}
As shown in Fig.~\ref{fig:worldspace}, despite that CoMoVi generates in-place motion, it can be easily extended to world space via a kinematic post-processing. Details are provided in the supplementary materials.

\begin{table*}[t!]
    \centering
    \scriptsize
    \setlength{\tabcolsep}{11.2pt}
    \begin{tabular}{lccccc}
        \toprule
        Method & SC $\uparrow$ & BC $\uparrow$ & MS $\uparrow$ & AQ $\uparrow$ & IQ $\uparrow$\\
        \midrule
        CogVideoX1.5-I2V-5B~\cite{yang2024cogvideox} & 0.933 & 0.947 & 0.991 & 0.512 & \underline{0.716}\\
        Wan2.2-I2V-5B~\cite{wan2025wan} & \underline{0.948} & \underline{0.960} & \underline{0.993} & \underline{0.529} & 0.713\\
        Go-to-Zero-7B+Champ~\cite{fan2025gotozero,zhu2024champ} & 0.931 & 0.934 & 0.980 & 0.479 & 0.663\\
        \cellcolor{lightgray!30}Ours & \cellcolor{lightgray!30}\textbf{0.955} & \cellcolor{lightgray!30}\textbf{0.963} & \cellcolor{lightgray!30}\textbf{0.993} & \cellcolor{lightgray!30}\textbf{0.533}& \cellcolor{lightgray!30}\textbf{0.718}\\
        \bottomrule
  \end{tabular}
  \vspace{2pt}
  \caption{Quantitative evaluation of human video generation using VBench metrics.}
  \label{tab:vigen}
\vspace{-25pt}
\end{table*}

\subsection{Comparisons on Video Generation Task}
\label{sec:comp_vigen}

\paragraph{\bf Baselines.}
We choose to compare with two leading open-source I2V models, CogVideoX1.5-I2V-5B~\cite{yang2024cogvideox} and Wan2.2-I2V-5B~\cite{wan2025wan}, to which the model size of CoMoVi is comparable. Additionally, we construct a cascaded baseline composed of SoTA text-driven motion synthesis~\cite{fan2025gotozero} and motion-driven video generation~\cite{zhu2024champ} to ensure a fair and meaningful comparison between different models and paradigms in similar scale and capability.

\vspace{-2mm}
\paragraph{\bf Results.} As depicted in Fig.~\ref{fig:vigen_comp}, CoMoVi gets benefits from our 2D motion representation, which encapsulates rich 3D motion information, generating realistic human videos with more consistent body structure, higher prompt fidelity, and anatomically plausible motions. While CogVideoX1.5~\cite{yang2024cogvideox} and Wan2.2-I2V-5B~\cite{wan2025wan} struggle with motion-prompt misalignment, distorted body shapes, and background maintenance. 
Though the motion-driven video generation methods achieve promising performances as shown in previous works~\cite{hu2024animate,zhu2024champ,wang2024humanvid}, cascading a SoTA T2M model Go-to-Zero~\cite{fan2025gotozero} and a SoTA motion-controllable video generation model Champ~\cite{zhu2024champ} leads to inferior results due to the limited generalization capability, low prompt fidelity, and misalignment with the video frames to guide video generation. Details are provided in the supplementary material. In contrast, CoMoVi achieves high-quality video generation by the co-generation of aligned motion and videos, avoiding reliance on external video and motion reference. 

The quantitative results detailed in Tab.~\ref{tab:vigen} demonstrate the advantage of our co-generative framework in all evaluation dimensions defined by VBench~\cite{huang2023vbench,huang2024vbench++,zheng2025vbench2}. 
It is noteworthy that VBench metrics~\cite{huang2023vbench,zheng2025vbench2,huang2024vbench++} prioritize consistency of subject identity and background, smoothness of pixel flow, and imagery clarity, but do not fully show the human motion quality and plausibility. As evidenced by Fig.~\ref{fig:vigen_comp} and Tab.~\ref{tab:vigen}, our method significantly outperforms baselines in body structure consistency and prompt fidelity, but the metric differences are within $\pm 0.05$ due to no significant variation or mutation exhibited in baseline results.

\subsection{Ablation Study}
\label{sec:ablation_study}

We further conduct extensive ablation studies to validate the significance of our 2D motion representation and the effectiveness of our model architecture design.

\vspace{-2mm}
\paragraph{\bf Ablation of 2D Motion Representations.} As introduced in Sec.~\ref{sec:morep}, our 2D human motion representation integrates surface normals with body part semantics. Therefore, we experiment with ``normal only'' and ``body semantic only'' settings, and also a commonly used 2D pose representation ``DWPose''~\cite{yang2023dwpose}. Tab.~\ref{tab:ablation} shows that directly fine-tuning Wan2.2-I2V-5B~\cite{wan2025wan} on RGB videos only without our co-generative framework (``w/o motion'') leads to significant performance degradation. While the absence of any factor among normals, body semantics, and surface rendering results in suboptimal performance in both generation tasks.

\begin{figure*}[t!]
    \centering
    \includegraphics[width=\linewidth]{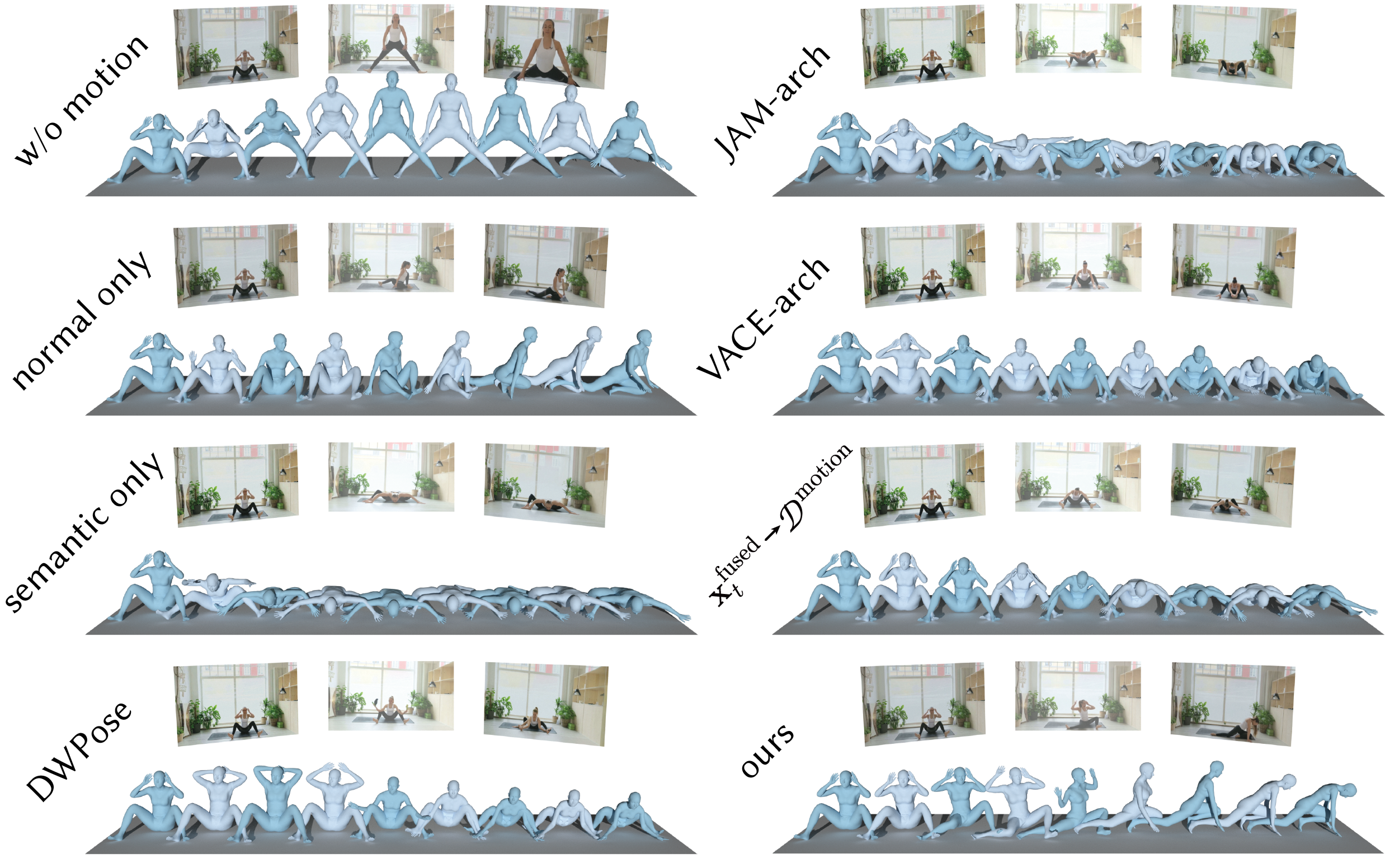}
    \vspace{-20pt}
    \caption{Qualitative results of different motion representations and model architectures. The input motion keyword is: \textit{``transition from seated state to get up and stretch body"}.}
    \label{fig:ablation}
\vspace{-5pt}
\end{figure*}

\vspace{-2mm}
\paragraph{\bf Ablation of Model Architecture Designs.} In designing our model architecture, we explore several approaches inspired by VideoJAM~\cite{chefer2025videojam} and VACE~\cite{jiang2025vace}, ultimately adopting a dual-branch diffusion model with the full copy strategy.

\vspace{-2mm}
\paragraph{Dual-Branch Architecture.} Specifically, we first refer to VideoJAM~\cite{chefer2025videojam} to concatenate RGB videos and our 2D motion representations along the latent channel dimension, and double the dimension of the patch embedding layer and the output head layer of Wan2.2-I2V-5B~\cite{wan2025wan}. However, we observe that this severely disrupts the pre-trained VDM's latent space during the initial training phase, requiring a computationally expensive reconstruction process. Consequently, this method fails to outperform baseline models with a limited training budget.

\begin{table*}[t!]
    \centering
    \scriptsize
    \setlength{\tabcolsep}{2pt}
    \begin{tabular}{lccccccccc}
        \toprule
        & \multicolumn{4}{c}{Motion Generation} & \multicolumn{5}{c}{Video Generation}\\
        \cmidrule(lr){2-5}\cmidrule(lr){6-10}
        Ablation Configs & FID $\downarrow$ & R@1 $\uparrow$ & R@3 $\uparrow$ & MMDist $\downarrow$ & SC $\uparrow$ & BC $\uparrow$ & MS $\uparrow$ & AQ $\uparrow$ & IQ $\uparrow$ \\
        \midrule
        \cellcolor{lightgray!30}Motion Representation & \cellcolor{lightgray!30} & \cellcolor{lightgray!30} & \cellcolor{lightgray!30} & \cellcolor{lightgray!30} & \cellcolor{lightgray!30} & \cellcolor{lightgray!30} & \cellcolor{lightgray!30} & \cellcolor{lightgray!30} & \cellcolor{lightgray!30} \\
        w/o motion & 0.758 & 0.458 & 0.556 & 4.079 & 0.937 & 0.944 & 0.980 & 0.520 & 0.709\\
        Normal only & \underline{0.415} & \underline{0.538} & 0.556 & \textbf{3.003} & 0.918 & 0.946 & 0.988 & 0.521 & 0.710\\
        Body semantic only & 0.553 & 0.499 & \underline{0.589} & 3.536 & 0.940 & 0.944 & \textbf{0.994} & 0.528 & 0.708\\
        DWPose & 0.503 & 0.464 & 0.551 & 3.396 & 0.929 & 0.951 & 0.981 & 0.522 & 0.701\\
        \midrule
        \cellcolor{lightgray!30}Architecture & \cellcolor{lightgray!30} & \cellcolor{lightgray!30} & \cellcolor{lightgray!30} & \cellcolor{lightgray!30} & \cellcolor{lightgray!30} & \cellcolor{lightgray!30} & \cellcolor{lightgray!30} & \cellcolor{lightgray!30} & \cellcolor{lightgray!30} \\
        VideoJAM joint latent space & 0.755 & 0.459 & 0.559 & 3.534 & 0.868 & 0.942 & 0.987 & 0.529 & 0.705\\
        VACE distributed copy & 0.625 & 0.500 & 0.572 & 3.148 & 0.927 & 0.949 & 0.994 & 0.522 & 0.715\\
        Pass $\bm{x}_t^{\text{fused}}$ to $\mathcal{D}^{\text{motion}}$ & 0.718 & 0.466 & 0.552 & 4.157 & 0.911 & 0.955 & 0.985 & 0.530 & \underline{0.718}\\
        Ours w/o $\mathcal{L}^{\text{smpl}}$ & - & - & - & - & \underline{0.951} & \underline{0.961} & 0.993 & \underline{0.532} & 0.714\\
        Ours w/o $\mathcal{D}^{\text{motion}}$ & - & - & - & - & 0.949 & 0.940 & 0.988 & 0.517 & 0.709\\
        \midrule
        \cellcolor{lightgray!30}Ours & \cellcolor{lightgray!30}\textbf{0.349} & \cellcolor{lightgray!30}\textbf{0.565} & \cellcolor{lightgray!30}\textbf{0.640} & \cellcolor{lightgray!30}\underline{3.035} & \cellcolor{lightgray!30}\textbf{0.955} & \cellcolor{lightgray!30}\textbf{0.963} & \cellcolor{lightgray!30}\underline{0.993} & \cellcolor{lightgray!30}\textbf{0.533} & \cellcolor{lightgray!30}\textbf{0.718}\\
        \bottomrule
    \end{tabular}
    \vspace{2pt}
    \caption{Quantitative ablation of representations and architectures.}
    \label{tab:ablation}
\vspace{-25pt}
\end{table*}

\vspace{-2mm}
\paragraph{Full Copy of DiT Blocks.} Therefore, we adopt dual-branch diffusion architecture following VACE~\cite{jiang2025vace}, which effectively preserves the integrity of the latent space of pre-trained VDM. Nevertheless, we identify that the distributed copy strategy proposed by VACE~\cite{jiang2025vace} causes the copied branch to completely lose the prior knowledge of pre-trained VDM. At the first training step, the output of the copied branch is totally noised, indicating that it essentially discards priors and requires to be trained from scratch. The quantitative performance shown in Tab.~\ref{tab:ablation} demonstrates that our dual-branch diffusion architecture with the full copy strategy achieves better performance in both generation tasks.

\vspace{-2mm}
\paragraph{Inject Information from $\mathcal{D}^{\text{rgb}}$ to $\mathcal{D}^{\text{motion}}$.} Furthermore, we experiment with passing the fused latent feature $\bm{x}_t^{\text{fused}}$ defined in Eq.~\ref{eq:fuse} rather than $\bm{x}_t^{\text{motion}}$ to the 2D motion diffusion branch $\mathcal{D}^{\text{motion}}$. Intriguingly, it is found that the direct feature injection from RGB to 2D motion latents significantly disturbs the denoising process of $\mathcal{D}^{\text{motion}}$. We analyze that 2D motion representation is far more sparse than RGB videos containing rich appearances and background details, which results in chaotic and fluctuating artifacts in 2D motion generation, degrading the preservation of essential 3D information. And eventually, the effectiveness of motion guidance is weakened, leading to the suboptimal performance (see Fig.~\ref{fig:ablation}).

\vspace{-2mm}
\paragraph{3D Regularization $\mathcal{L}^{\text{smpl}}$.} As stated in Sec.~\ref{sec:comovi}, the additional 3D regularization term $\mathcal{L}^{\text{smpl}}$ is introduced to reinforce the alignment between 2D latents and 3D representation, ensuring structure consistency and visual plausibility of generated human videos. Without $\mathcal{L}^{\text{smpl}}$, as we quantitatively evaluate in Tab.~\ref{tab:ablation}, the quality of video generation results degrade especially in ``Subject Consistency'' dimension.

\vspace{-2mm}
\paragraph{Directly Use 3D Motion without $\mathcal{D}^{\text{motion}}$.} Finally, we validate the importance of our 2D motion branch $\mathcal{D}^{\text{motion}}$. Tab.~\ref{tab:ablation} demonstrates that discarding our 2D motion representation and $\mathcal{D}^{\text{motion}}$ causes performance degradation in all five evaluation dimensions on video generation task. We analyze that, different from motion-driven video generation methods like Animate Anyone~\cite{hu2024animate} and Champ~\cite{zhu2024champ} which require a reference video and pre-extracted 2D pose or 3D motion sequence from it, co-generation of human motion and video necessiates a tight coupling the generation process of the two modalities to ensure the alignment of gerated 3D motions and 2D videos. Consequently, our 2D motion representation is not only pixel-aligned with RGB video, but also encodes rich 3D structure and semantic information, which is incorporated in a shared denoising process with VDM via $\mathcal{D}^{\text{motion}}$. This design thus achieves superior performance to directly using 3D motion conditioning without $\mathcal{D}^{\text{motion}}$.

%% file: sec/5-conclusion.tex
\section{Conclusion}
\label{sec:conclusion}
In this work, we propose a novel framework called CoMoVi for co-generation of 3D human motions and realistic videos. Our key idea is to couple the denoising process of 3D motions and 2D videos, enabling synchronous generation of both within a single diffusion loop. We first introduce a novel 2D motion representation that encodes surface normals and body part semantics of 3D SMPL mesh into RGB images, allowing it to directly inherit priors from pre-trained VDMs. Then, we develop a dual-branch diffusion model with mutual feature interactions and 3D-2D cross-attentions, providing motion guidance for video generation while propagating VDM's generalization capability to 3D motion generation. Moreover, we contribute the CoMoVi-Dataset, a large-scale human video collection annotated with high-quality text and motion labels to support versatile video-based and motion-related tasks. Comprehensive experiments on multiple benchmarks demonstrate our method's effectiveness in both motion and video generation.

%% file: sec/X_suppl.tex
\setcounter{page}{1}

\appendix
\setcounter{figure}{0}
\setcounter{table}{0}

\section{More Implementation Details}
\label{sec:more_implementation_detail}
\begin{wraptable}{r}{0.48\textwidth}
    \centering
    \scriptsize
    \setlength{\tabcolsep}{0.01pt}
    \vspace{-20pt}
    \begin{tabular}{lc}
        \toprule
        Hyper-parameter & Value\\
        \midrule
        Batch Size / GPU & 1 / 24\\
        Accumulate Step & 4\\
        Optimizer & AdamW\\
        Weight Decay & 3e-2\\
        Learning Rate & 2e-5\\
        AdamW $\beta_1$ & 0.9\\
        AdamW $\beta_2$ & 0.999\\
        Scheduler & Constant with Warmup\\
        Warmup Steps & 100\\
        Training Steps & 6,000\\
        Resolution & $704\times 1280$\\
        Max Gradient Norm & 0.5 \\
        Weighting Scheme & Uniform\\
        Pre-trained Model & Wan2.2-I2V-5B\\
        \midrule
        Sampler & Flow Unipc\\
        Sample Steps & 50\\
        CFG Scale & 6.0\\
        \midrule
        Training Device & A100-SXM4-40G$\times$24\\
        Training Strategy & ZeRO-3\\
        Gradient Checkpointing & True\\
        CPU Offload & True\\
        Precision & bf16\\
        \midrule
        Training Speed & 280s per accumulation\\
        Inference Speed & 15min per video in 5s\\
        \bottomrule
    \end{tabular}
    \vspace{-2pt}
    \caption{Hyper-parameters of our model training and inference.}
    \label{tab:hyperparam}
    \vspace{-10pt}
\end{wraptable}

The main paper outlines the primary training and experimental parameters. This section provides a comprehensive description of multiple stages of our model training, and the detailed procedures of the comparative experiments on motion and video generation tasks.

\vspace{-2mm}
\paragraph{\bf Training.}
Our model training is conducted in two stages. The first stage adapts the copied motion DiT branch $\mathcal{D}^{\text{motion}}$ to our 2D motion representation domain using $\mathcal{L}^{\text{motion}}$ only. During this stage, only $\mathcal{D}^{\text{motion}}$ is enabled with gradient calculation and weight update, trained for 2,000 steps. In the second stage, we incorporate the mutually interactive zero-linear layers and 3D-2D cross-attention modules into the training process, using the total loss $\mathcal{L}^{\text{total}}$ to supervise it. To enhance training efficiency and reduce GPU memory consumption, at each training step we randomly select the latent features from three DiT layers, while consistently including the final layer of $\mathcal{D}^{\text{video}}$ and $\mathcal{D}^{\text{motion}}$, to operate feature fusion and perform cross-attention with the 3D motion query. The pre-trained weights of RGB DiT branch $\mathcal{D}^{\text{video}}$ are frozen, while the remaining components are updated over 4,000 steps at the second training stage. In total, the model is trained for 6,000 steps across both stages. Detailed hyperparameter configurations are provided in Tab.~\ref{tab:hyperparam}.

\begin{figure*}[t!]
    \centering
    \includegraphics[width=\textwidth]{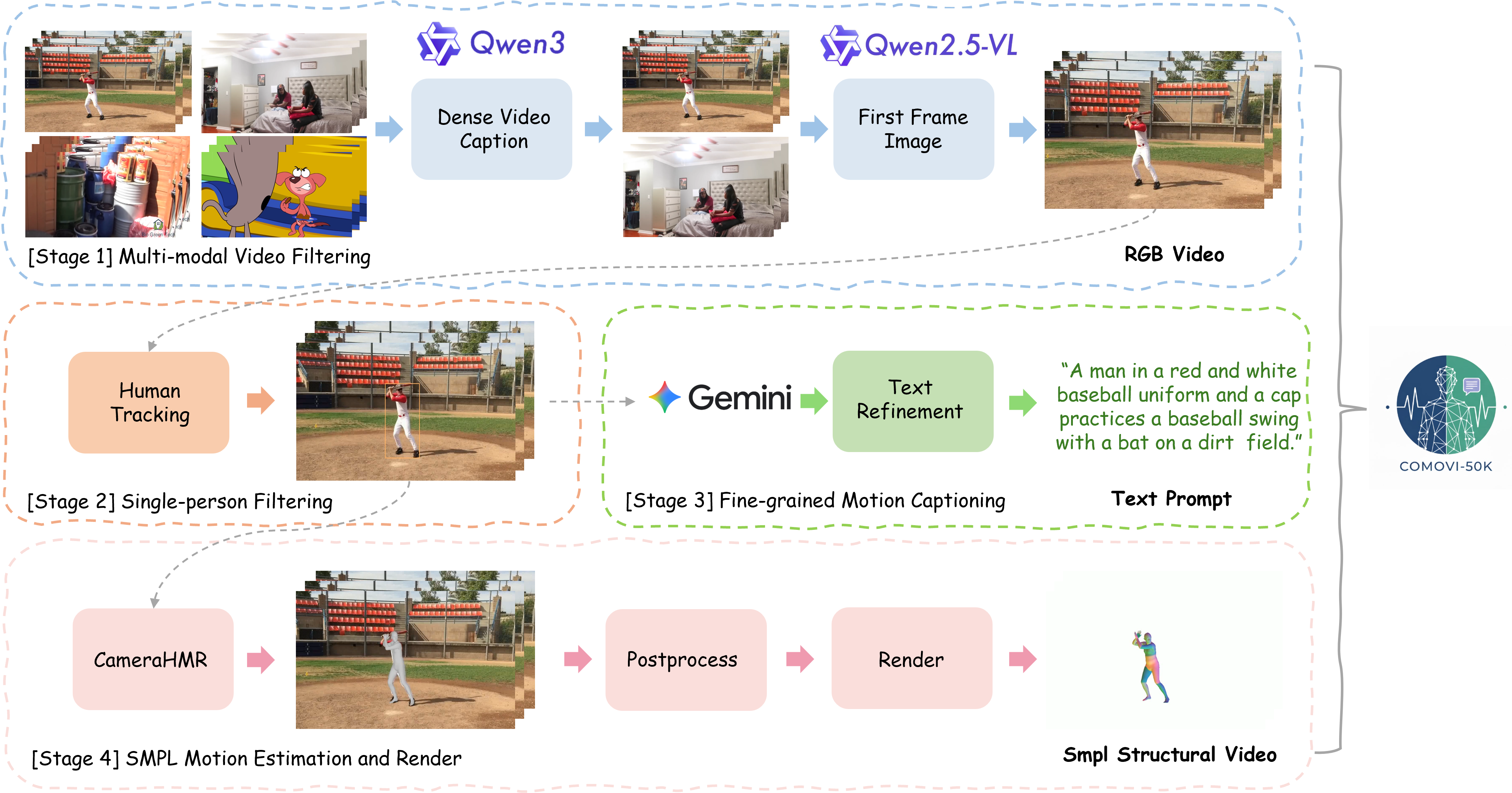}
    \vspace{-20pt}
    \caption{Curation pipeline of our CoMoVi-Dataset.}
    \label{fig:data_curation}
\vspace{-15pt}

\end{figure*}

\vspace{-2mm}
\paragraph{\bf Experiments of Motion Generation.}
For 3D human motion evaluation, we follow the 263-dimensional representation and use the pretrained motion and text encoder provided by HumanML3D~\cite{guo2022ml3d}. The metrics reported in the main paper are determined by the average of 20 independent inference results with a confidence interval of 95\% for each model.

\begin{figure*}
    \begin{widecodebox}
        \texttt{\textbf{INPUT:}}\\
        A dense caption \texttt{cap} of the input video.\\
        \\
        \texttt{\textbf{PROMPT:}}\\
        ``\\
        Read and understand the given caption of a video \{\texttt{cap}\}, then answer the following questions:\\
        - Does the video capture humans?\\
        - Does the video capture only one person?\\
        - Does the video capture the whole body of the person all the time?\\
        - Does the video capture interactions behavior between the person and surroundings, like objects or the scene?\\
        \\
        You MUST answer and only answer the given questions by saying `YES' or `NO'.\\
        You MUST return your output in a list format. Here is an example: [`YES', `YES', `NO', `NO'].\\
        "\\
        \\
        \texttt{\textbf{OUTPUT:}}\\
        \text{[}``answer for question 1", ``answer for question 2", ``answer for question 3", ``answer for question 4"\text{]}
    \end{widecodebox}
    \vspace{-10pt}
    \caption{Prompt instruction for Qwen3~\cite{qwen3} to analyze dense video captions.}
\label{fig:qwen3_prompt}
\vspace{-15pt}
\end{figure*}

\begin{figure*}
    \begin{widecodebox}
        \texttt{\textbf{INPUT:}}\\
        The first frame \texttt{frame} of the input video.\\
        \\
        \texttt{\textbf{PROMPT:}}\\
        ``\\
        You will be given the image \{\texttt{frame}\}, answer `YES' if and only if the following conditions are met:\\
        - There exists one and only one human in this image.\\
        - The human is not a child.\\
        - The whole body of the human is visible in the image, from head to feet.\\
        - The human is not giving a talk, interview, presentation or speech.\\
        - The image is not a game, television, movie, anime or cartoon screen.\\
        Otherwise, answer `NO'.\\
        \\
        Remember, you MUST only answer `YES' or `NO'.\\
        "\\
        \\
        \texttt{\textbf{OUTPUT:}}\\
        `YES' or `NO'
    \end{widecodebox}
    \vspace{-10pt}
    \caption{Prompt instruction for Qwen2.5-VL~\cite{qwen3} to analyze the first frame of video.}
\label{fig:qwenvl_prompt}
\end{figure*}

\begin{figure*}
    \begin{widecodebox}
        \texttt{\textbf{INPUT:}}\\
        A video \texttt{video}.\\
        \\
        \texttt{\textbf{PROMPT:}}\\
        ``\\
        You will be shown a human video \{\texttt{video}\}. You should identify the most prominent subject in this video and describe the appearance and motion of that person only, ignore other people. You should also describe the objects interacting with or near to the human if any. Don't use plural words like `they/their/them', start your description with `A/The person/man/male/boy/woman/female/girl'. If you use the word `left' or `right', `forward' or `backward', you should always describe it relative to the person's body, not the camera or screen.\\
        \\
        You are required to return a short caption of \{\texttt{video}\}. Give a brief and accurate description of the human appearance, clothing, and motion in only one sentence. Describe the appearance and clothing first and then the motion with possible interactions with surronding objects. Use the specific and proper terms to describe the motion whenever possible.\\
        "\\
        \texttt{\textbf{OUTPUT:}}\\
        A short caption.
    \end{widecodebox}
    \vspace{-10pt}
    \caption{Prompt instruction for Gemini2.5-Pro to caption human motion in videos.}
\label{fig:gemini_prompt}
\vspace{-15pt}
\end{figure*}

\vspace{-2mm}
\paragraph{\bf Experiments of Video Generation.}
For fair comparative experiments of video generation task, all models are evaluated at their optimal resolutions ($768\times 1360$ for CogVideoX1.5-I2V-5B~\cite{yang2024cogvideox}, and $704\times 1280$ for Wan2.2-I2V-5B~\cite{wan2025wan} and ours), and each model is configured to generate video sequences of a unified length of 81 frames. All results are generated once using random seed 42 and CFG scale 6.0 without any cherry-picking.

\section{CoMoVi-Dataset Curation}
\label{sec:dataset_curation}
In the main paper, we give an overview of our dataset construction process. Here, we elaborate on the detailed procedures. As illustrated in Fig.~\ref{fig:data_curation}, firstly, we curate a collection of high-quality videos sourced from the Internet and the Koala-36M dataset~\cite{wang2025koala}, and employ LLMs/VLMs~\cite{qwen3,qwen2.5} to selectively retain only those depicting single-person movements (Sec.~\ref{sec:stage1}). Subsequently, we apply 2D human pose detection model~\cite{redmon2016yolo,xu2022vitpose} to filter out videos where the subject is largely outside the frame or severely occluded (Sec.~\ref{sec:stage2}). The filtered videos are finally captioned (Sec.~\ref{sec:stage3}) and labeled with 3D motion (Sec.~\ref{sec:stage4}). We provide specific prompt instructions used at each filtering stage and our confirmation of dataset ethics (Sec.~\ref{sec:data_ethics}) below.

\subsection{Multimodal Video Filtering}
\label{sec:stage1}

Given a large collection of publicly available videos, we ask the Qwen3 model~\cite{qwen3} to determine if the video contents satisfy our criteria according to the corresponding dense captions using the prompt instruction shown in Fig.~\ref{fig:qwen3_prompt}. We only keep the videos that are judged as ``\text{[}`Yes', `Yes', `Yes', `Yes'\text{]}" in this initial text-based filtering stage, and then process them using the Qwen2.5-VL model~\cite{qwen2.5}, which takes the first frame of each video and the prompt instruction specified in Fig.~\ref{fig:qwenvl_prompt} to make a visual confirmation. This multimodal filtering procedure effectively filters out non-human videos, multi-person videos, and contents such as movies, animations, and video games that may contain human-like characters but do not represent real-world human motions.

\subsection{Human Tracking Filtering}
\label{sec:stage2}

To ensure a balanced data distribution and avoid overwhelming clips from long videos, we segment each video into non-overlapping 5-second clips, with a maximum of two clips retained per video. Subsequently, we apply human tracking utilizing YOLO~\cite{redmon2016yolo} and ViTPose~\cite{xu2022vitpose} for each clip. A series of confidence thresholds are established to flag frames with low-confidence human detections throughout a sequence. Clips containing an excessive number of such low-confidence frames are filtered out to ensure data quality.

\subsection{Video Captioning}
\label{sec:stage3}

For all retained 5-second clips, the human motion captioning is performed by querying the Gemini-2.5-Pro API at 1 fps, adhering to the official recommendation. The prompt instruction provided for motion captioning is shown in Fig.~\ref{fig:gemini_prompt}. We also experiment with higher frame rates, which do not yield improvement in caption quality but substantially increase the cost of annotation.

\subsection{3D Human Motion Annotation}
\label{sec:stage4}

We estimate the SMPL parameters~\cite{loper2023smpl} for each frame of the filtered human videos using CameraHMR~\cite{patel2025camerahmr}. Since these per-frame estimations are independent, which can lead to motion jitter due to occlusions or motion blur in videos, we apply the smoothing curve in Blender to post-process the estimated body motions to ensure temporal coherence. For consistency of body shape across the video sequence, the SMPL shape parameters estimated from the first frame are applied to all subsequent frames.

\subsection{Dataset Ethics}
\label{sec:data_ethics}
We strictly adhere to the conference ethics guidelines and only collect publicly available videos from academic datasets and open social media platforms. We confirm that all collected data are used only for research purposes. This condition will also be explicitly stated upon the release of our dataset. We fully respect the privacy of the individuals appearing in the videos, so no personal information or metadata is retained. Furthermore, in compliance with video ownership rights, only video identifiers will be publicly released, rather than the original video files.

\section{More Discussions}

\subsection{How to Extend CoMoVi Results to World-Space}
As shown in Fig. 7 in the main paper, a simple and deterministic kinematic approach can be employed to ``lift'' in-place motion generated by CoMoVi to world space via physical constraints of locomotion that leverages the zero-velocity assumption of the stance foot. Specifically, for each time step $t$, we identify the stance joint $k$ (e.g., the ankle or toe) as the end-effector exhibiting the minimal vertical elevation relative to the root. Assuming the stance foot remains static in the world coordinate system (i.e., $\mathbf{v}_{k}^{\text{world}} \approx 0$), the global linear velocity of the root joint, $\mathbf{v}_{\text{root}}$, can be derived by inverting the relative velocity of the stance foot: $\mathbf{v}_{\text{root}} \approx - \dot{\mathbf{p}}_{k}^{\text{local}}$, where $\mathbf{p}_{k}^{\text{local}}$ denotes the position of the stance foot in the root-relative coordinate system. The final global trajectory is obtained via temporal integration of these frame-wise velocities
\begin{equation}
\begin{aligned}
    \mathbf{v}_{\text{root}}^{(t)} \approx - \dot{\mathbf{p}}_{k}^{\text{local}} &= - \left( \mathbf{p}_{k}^{(t)} - \mathbf{p}_{k}^{(t-1)} \right) / \Delta t\\
    \mathbf{T}^{(t)} &= \mathbf{T}^{(t-1)} + \mathbf{R}^{(t-1)} \mathbf{v}_{\text{root}}^{(t)} \Delta t,
\end{aligned}
\end{equation}
where $\mathbf{T}$ and $\mathbf{R}$ represents the root translation and orientation respectively.

\subsection{Denoising Guidance of 3D Motion during Inference}
\begin{figure*}[t!]
    \centering
    \includegraphics[width=\textwidth]{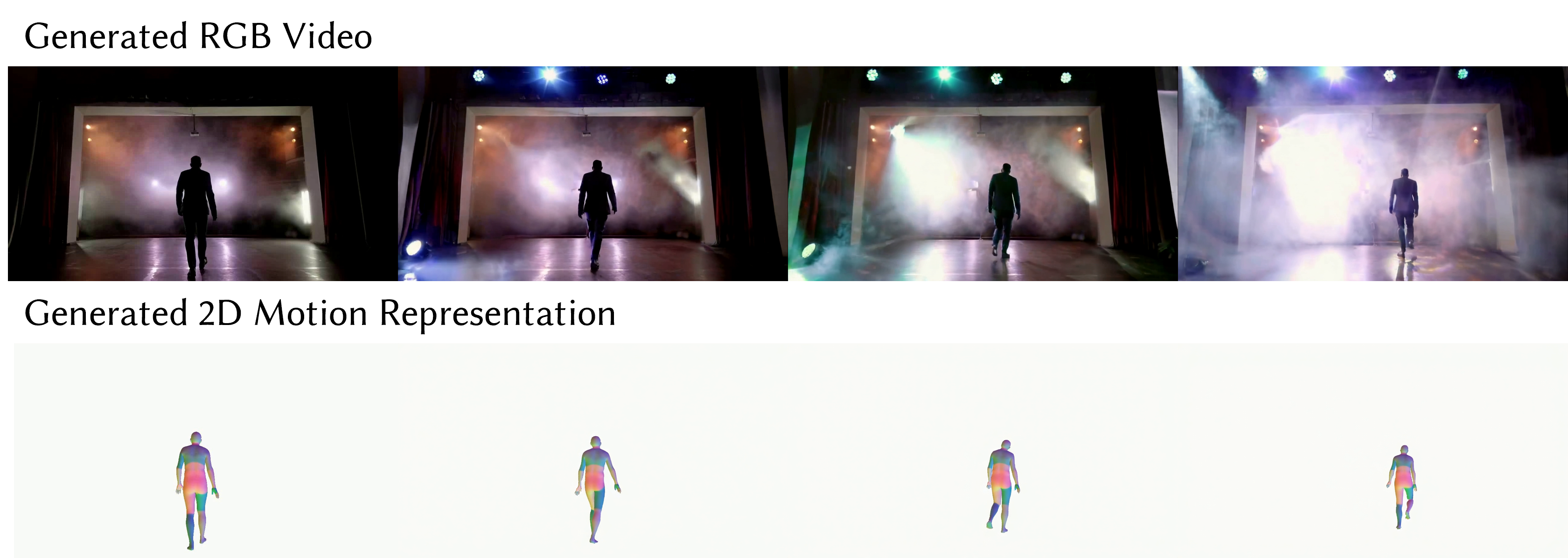}
    \vspace{-20pt}
    \caption{Synchronous and pixel-aligned generation of RGB video and our 3D-aware 2D motion representation sequence.}
    \label{fig:2d_motion_rgb}
\vspace{-15pt}
\end{figure*}

During inference, the 3D guidance is provided by our 2D motion representation (see Fig.~\ref{fig:2d_motion_rgb}), which is 3D-aware (containing both body surface normals and semantics information) and pixel-aligned with the RGB video. We introduce the 3D-2D cross-attention module to learn to output 3D motion from DiT latents, achieving synchronous geneartion of 3D motion and 2D video. This design is also inspired by REPA~\cite{yu2024repa}, regularizing the diffusion training process via additional 3D regularization $\mathcal{L}^{\text{smpl}}$.

\subsection{Evaluating on Motion-X++ Benchmark}
In the field of text-driven human motion generation, both HumanML3D~\cite{guo2022ml3d} and Motion-X++~\cite{lin2023motionx} are commonly used benchmarks. However, HumanML3D~\cite{guo2022ml3d} focuses solely on text-motion pairs, whereas Motion-X++~\cite{lin2023motionx} is a tri-modal dataset of text, image, and 3D human motion, which aligns the task investigated in this work. Moreover, the research problem of this work is co-generating 3D human motion and \textbf{realistic} video, which means synthetic data is undesired and \textbf{real-world} video data is required. We therefore adopt the Motion-X++ benchmark~\cite{lin2023motionx} for our evaluation of the motion generation task.

\section{Limitations and Future Work}
While our method offers significant advantages, it is constrained to generating fixed-length motion sequences, lacking the capacity for variable-length or infinite-length generation. Additionally, due to the inherently denser nature of video latents compared to 3D human motion data, our inference speed is relatively slower. Promising future directions include extending our framework to human-object interaction scenarios, employing distillation techniques for generation acceleration, and enabling the generation of variable-length sequences.